# Fast and robust pushbroom hyperspectral imaging via DMD-based scanning

Reza Arablouei, Ethan Goan, Stephen Gensemer, and Branislav Kusy
Commonwealth Scientific and Industrial Research Organisation, Pullenvale QLD, Australia

**ABSTRACT**

We describe a new pushbroom hyperspectral imaging device that has no macro moving part. The main components of the proposed hyperspectral imager are a digital micromirror device (DMD), a CMOS image sensor with no filter as the spectral sensor, a CMOS color (RGB) image sensor as the auxiliary image sensor, and a diffraction grating. Using the image sensor pair, the device can simultaneously capture hyperspectral data as well as RGB images of the scene. The RGB images captured by the auxiliary image sensor can facilitate geometric co-registration of the hyperspectral image slices captured by the spectral sensor. In addition, the information discernible from the RGB images can lead to capturing the spectral data of only the regions of interest within the scene. The proposed hyperspectral imaging architecture is cost-effective, fast, and robust. It also enables a trade-off between resolution and speed. We have built an initial prototype based on the proposed design. The prototype can capture a hyperspectral image datacube with a spatial resolution of 400×400 pixels and a spectral resolution of 500 bands in less than thirty seconds.

## 1. INTRODUCTION

Hyperspectral imaging is a combination of spectroscopy and image forming. It creates an electromagnetic profile of a two-dimensional (2D) scene by capturing the intensity of radiation emitted from each elementary spatial segment of the scene (pixel) at various wavelengths. The collected information is usually arranged into a hyperspectral image datacube, which has two spatial dimensions and one spectral dimension. The datacube can be seen as a batch of monochromatic images each corresponding to one spectral band. Hyperspectral imaging systems divide the radiation received from each pixel into many sections, i.e., narrow spectral bands often only a few nanometers apart, before detecting them. Therefore, in these systems, the number of photons detectable at each wavelength is greatly reduced hence the required sensor sensitivity or exposure time is increased in comparison with the conventional panchromatic or trichromatic imagers.

The spatial and spectral information provided by a hyperspectral image of a scene can be analyzed to detect, identify, or discriminate objects and patterns as well as the chemical composition of the material present at the scene. Hyperspectral imaging finds application in many areas such as computer vision, remote sensing, biomedicine, surveillance, precision agriculture, environmental studies, forensics, nanoparticle research, food science, mining, forestry, etc. [1]-[5].

Several techniques have been developed for hyperspectral imaging. They can roughly be categorized as whiskbroom (point scan), pushbroom (line scan), tunable filter (wavelength scan), and snapshot.

The whiskbroom technique is mostly popular in remote sensing applications. It is based on spectroscopy of a single pixel at a time. Whiskbroom hyperspectral imagers are usually mounted on flying platforms such as aircrafts or satellites and employ a rotating mirror to sweep a scan line perpendicular to the motion direction of the platform. With the pushbroom technique, spectra of the pixels on a single spatial line are sensed at each moment in time. Therefore, a two-dimensional array of measurements is created for each spatial line. One of the dimensions of the array corresponds to the spatial axis along the sensed line and the other corresponds to the spectral axis. While stepping across the scene in a scan direction perpendicular to the axis of sensed lines, multiple 2D arrays are collected. These arrays are stacked together to form a hyperspectral image datacube. Scanning the scene is typically realized by moving the imager or its field of view across the scene. The speed of the movement should to be synchronized with the rate of capture in the imager [6]-[8]. Figure 1 depicts an example of spatial-spectral arrays captured by a pushbroom hyperspectral imager as well as their assembly within a hyperspectral image datacube.

Some hyperspectral imaging systems utilizes a tunable filter such as liquid crystal tunable filter (LCTF) or acousto-optical tunable filter (AOTF) to form a hyperspectral image of a scene by scanning across the desired spectral bands. They use the tunable filters to sense the entire scene in just one spectral band at a time. The LCTF uses the birefringence property of a liquid crystal to filter the desired spectral band while the AOTF uses a special crystal and acoustic waves to resolve the bands [9]-[12].

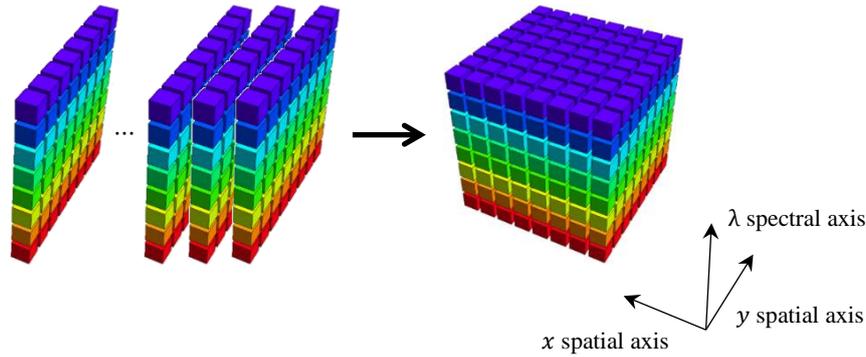

Figure 1, A hyperspectral image datacube and its spatial-spectral slices.

In all abovementioned techniques, different parts of the hyperspectral image datacube are sensed at consecutive points in time. Snapshot spectral imagers acquire the entire datacube at a single distinct point in time. They usually use a 2D image sensor and implement some technique to project the datacube onto the sensor [13]-[19]. The snapshot multispectral imager described in [13] has been developed by monolithically integrating a set of Fabry-Pérot interferometers [20], which are organized as pixel-level filter mosaics, on top of a standard off-the-shelf CMOS image sensor. Some snapshot hyperspectral imagers can be faster than their scanning-based counterparts. However, the limitation of pixel real-estate on image sensors imposes a trade-off between spatial and spectral resolution of the hyperspectral images captured by most snapshot spectral imagers.

Irrespective of the technique used for acquiring the image, most current hyperspectral imaging systems have inherent difficulties with georeferencing the captured hyperspectral image data. They often require extra processing to match the captured data with the geographical or topographical information of the scene or register images taken by sensors that have different (evenly though slightly) objective views of the scene [21]. In addition, most currently available hyperspectral imagers perform static data acquisition, which is to record the hyperspectral data of the whole of a given static scene with pre-set spatial and spectral resolutions. This typically makes them slow and inflexible hence unsuitable for applications where spectral information of only parts of a scene is needed. Furthermore, most hyperspectral imaging devices currently available on the commercial market are built in limited numbers for specific research or industrial applications. Therefore, they are generally not accessible to every curious mind on a budget.

Our objective is to build a low-cost hyperspectral imager that is fast, robust, and flexible to be utilized in various applications. In this paper, we present the design of a pushbroom hyperspectral imager that uses a digital micromirror device (DMD) for scanning the field of view. It contains two image sensors that can be triggered simultaneously. By design, the two image sensors have the same objective view of the scene and hence are perfectly aligned. We use one of the image sensors to capture the hyperspectral data of the scene and the other to take RGB images from the scene. The device can scan through the scene without the need for any moving part thanks to the use of the DMD. The RGB images captured along the hyperspectral data can aid in stitching the collected hyperspectral image slices together to form a hyperspectral image datacube. This feature together with the lack of any macro moving part endows the device with robustness against small movements or vibrations in the scene or the device. Another important advantage of the device is that, in any particular application, the RGB images of the scene taken by the respective image sensor can be examined to determine the parts of the scene that are of importance, e.g., leaves or crops in agriculture applications. By exploiting this information, the device can be instantly reprogrammed to capture the hyperspectral data only from the selected parts of the scene. This can dramatically increase the speed and efficiency of acquiring the hyperspectral data in many applications. In addition, the width of the effective aperture scanning the scene can be easily varied via DMD programming. This enables a trade-off between spatial/spectral resolution of the captured hyperspectral image datacube and the acquisition time.

We have materialized our design concept through building a prototype using off-the-shelf components encased within a 3D-printed enclosure. The prototype covers the visible near-infrared (VNIR) range with several hundred spectral bands separated by less than a nanometer. We present initial results obtained by the prototype and verify its accuracy by comparing with a commercially-renowned hyperspectral imaging device.

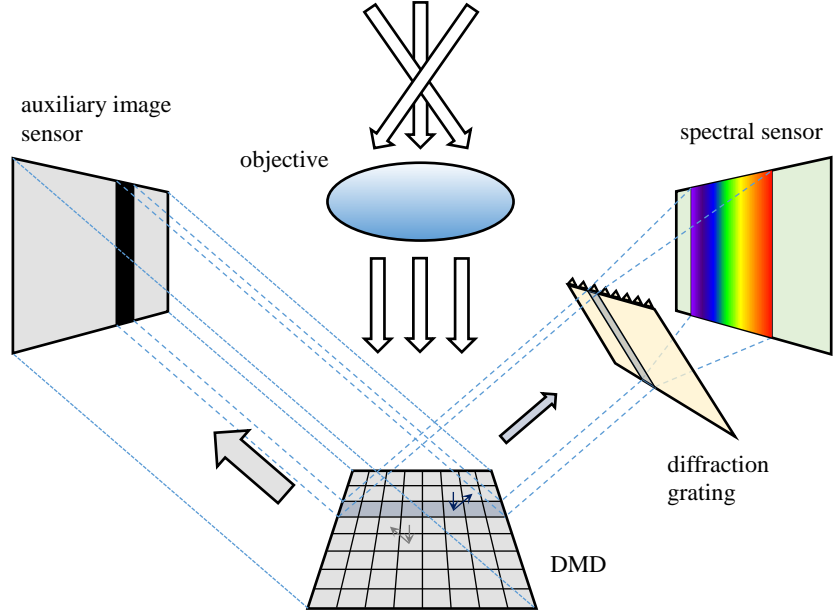

Figure 2, Illustration of the basic architecture of the presented hyperspectral imaging system. The dimensions are exaggerated for demonstrative purposes and are not to scale.

## 2. ARCHITECTURE

### 2.1. Overview

The developed device operates as a pushbroom hyperspectral imager. It constructs the hyperspectral image datacube of a 2D scene by capturing 2D slices of the 3D datacube in a sequential manner. Each slice is along the spectral dimension and one spatial dimension, e.g., vertical direction in the field of view. The slicing process can be seen as projecting the datacube onto the sensor plane sequentially in time.

The key components of the device are a DMD, a transmissive diffraction grating, and two 2D image sensor arrays, called the spectral sensor and the auxiliary image sensor. Figure 2 is an illustration of the basic architecture of the device. As depicted in the figure, the light reflected/emitted from the scene is collimated and focused onto the DMD. The DMD is programmed to successively switch through a set of non-overlapping single-line patterns that cover the whole or parts of the scene. The DMD effectively simulates a moving slit and, with each pattern, steers the light of a narrow section of the scene, e.g., a column or row of pixels, towards the diffraction grating and the rest of the light towards the auxiliary image sensor. The spectral components of the light of the narrow section are separated by the diffraction grating and captured by the spectral sensor as a slice of the hyperspectral image datacube of the scene. At the same time, the auxiliary image sensor, captures an RGB image of the scene containing a dark stripe that matches the position of the narrow section selected via the DMD pattern within the scene thus representing the spatial location of the corresponding hyperspectral slice. The DMD and the image sensors are synchronized using the available hardware input/output triggers.

Since the light reaching both image sensors comes from the same objective view and goes through the same collimating fore-optics, the two image sensors can be perfectly aligned without suffering from any parallax effect. Therefore, the RGB images captured by the auxiliary image sensor can serve two main purposes. First, they can facilitate the co-registration of the captured hyperspectral image slices against the scene. In other words, the information discernible from the RGB images can help decide which part of the scene each captured slice belongs to and how the slices need be arranged together to generate a hyperspectral image datacube, even at the presence of small movements or vibrations. Second, in applications where only the spectral data of only small parts of the scene is demanded, analyzing an initial RGB image of the scene captured by the auxiliary image sensor can determine those parts in the scene. Once determined, the spectral data collection can be restricted to the selected parts. This can result in a dramatic reduction in the acquisition time.

In the following, we describe the role and specifications of each key component of the new hyperspectral imaging system in more detail.

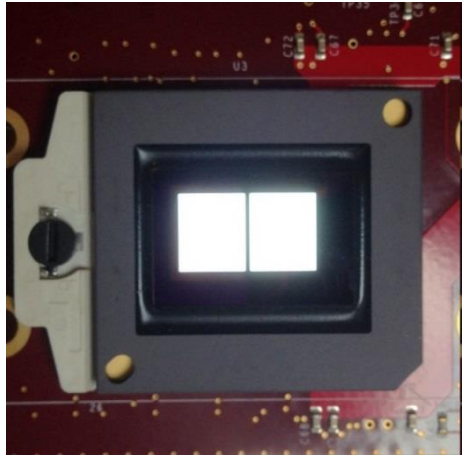

Figure 3, A DMD with a column of its micromirrors being in a different flip state from the rest of the micromirrors.

## 2.2. DMD

In conventional pushbroom hyperspectral imaging devices, the scene is swept across by a narrow moving slit often positioned perpendicular to the scan direction. During the scan, the light from the scene is blocked except for the part that passes through the slit, which is usually a narrow opening. In our system, we use a DMD to replace such a moving-slit mechanism. The main duty of the DMD is to select a narrow section of the scene and direct its light towards the diffraction grating and the spectral sensor. Hence, the DMD essentially functions as a moving slit while the only mechanically moving components are the DMD's micromirrors. Due to the micronic size of the micromirrors, they are highly resilient to fatigue and can be operated (flipped) at high rates of thousands of times per second. In addition, in conventional moving-slit designs, a large portion of the light that does not go through the slit is dissipated. In our DMD-based scanning design, we make use of the portion of the light that is not directed to the diffraction grating/spectral sensor by directing it to an auxiliary image sensor. Therefore, the DMD, the diffraction grating, and the image sensors are arranged in a way that each micromirror faces either the diffraction grating or the auxiliary image sensor at any given time depending on its flip state.

In most DMDs, the micromirrors have two possible flip states in which they tilt either $-12$ or $+12$ degrees relative to the horizon of the chipset substrate. We place the image sensors in our system according to these directions. We realize the swath scanning across the scene by loading the required patterns into the memory of the DMD driver. Once the memory is loaded with the desired patterns, the DMD driver can sequentially change the patterns. The speed of pattern change on the DMD can be accurately scheduled to account for the time duration required for image sensor exposure in any specific situation. Figure 3 displays a DMD where a column of its micromirrors have a different flip state from the rest of the micromirrors.

In our initial prototype, we have used a DLP6500FYE DMD together with a DLPC900 DMD digital controller supplied by Texas Instruments in their DLP® LightCrafter™ 6500 Evaluation Module. The DLP6500FYE has an array of 1920×1080 programmable micromirrors (true HD resolution). The DLPC900 enables flexible formatting and sequencing of arbitrary patterns to be realized by the micromirrors with rates up to 9,523 Hz. The evaluation module also offers a configurable input/output trigger for synchronization with any peripheral device [22].

We use a USB port to communicate with the DMD controller. We have developed the code for controlling the DMD in C programming language. It is designed to run on an embedded microprocessor using a Linux-based operation system. This allows the DMD to be controlled using light-weight and low-cost hardware/software. Although we have made our current code for the DLPC900 DMD driver, it can be easily modified to suit other DMD drivers. Our end goal is to develop a comprehensive library to facilitate programming of any DMD driver that admits communication via a USB port.

## 2.3. Image sensors

For the spectral and auxiliary image sensors, we use a pair of near-identical standard off-the-shelf CMOS image sensors. They have global shutter, dynamic range of 8 bits, resolution of up to 1280×960 pixels, pixel size of 3.75×3.75 $(\mu m)^2$, and frame rate of 25 to 60 frames per second. They also come with hardware input trigger and USB interface. The only difference between the two image sensors is that the auxiliary image sensor is overlaid by a Bayer filter array while the spectral sensor is not covered by any filter and operates as a monochromatic image sensor. Figure 4 shows the block diagram of the used image sensors.

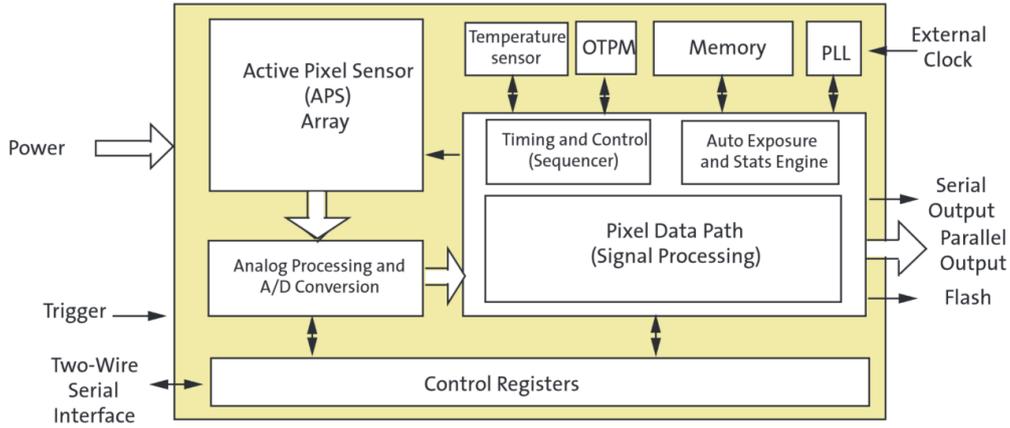

Figure 4, Block diagram of the used image sensors (from http://www.theimagingsource.com/).

We use the Video for Linux 2 (V4L2) framework to control the image sensors. V4L2 is a collection of device drivers and an API for supporting real-time video capture on Linux systems [23]. We utilize the hardware input/output triggers available for the DMD and the image sensors to synchronize them and adjust the exposure time of the image sensors in coordination with the DMD, which can send timed trigger signals upon changing the micromirror patterns. External hardware triggering enables adjustable and highly accurate timing. It also allows for both image sensors to operate and capture their respective data simultaneously.

### 2.4. Diffraction grating

Every pushbroom spectral imager required a dispersive element to split the light into its constituent spectral components. For this purpose, we use a transmissive diffraction grating. It is an optical piece with a periodic structure that produces several light beams with different wavelengths that travel in different directions. The deviation angle of each beam depends on its wavelength as well as the dispersion of the grating and the aperture width of the slit [24]. Consequently, the dispersion of the grating and the slit size determine the width of detected spectral bands and their corresponding areas on the spectral sensor. The wider the slit is, the lower the resolution in the spectral and one of the spatial dimensions is. However, a wider slit lets more light through allowing for a shorter exposure time and therefore faster data acquisition. The ability to adjust the slit width simply by programming the DMD allows us to negotiate a trade-off between resolution and speed without replacing or adjusting any optical component.

## 3. PROTOTYPE AND INITIAL RESULTS

### 3.1. Prototype

We have built a prototype based on the architecture described in Section 2. We fit the DMD and the image sensors as well as the diffraction grating and other optical components inside a 3D-printed enclosure shown in Figure 5. Figure 6 shows the prototype connected to the DMD controller board. The prototype captures hyperspectral images with spatial resolution up to 400×400 pixels. It covers a large part of the VNIR range, i.e., wavelengths between 400 and 900 nm, with up to 500 spectral bands each less than a nanometer wide. The exposure time of the image sensors is adjusted depending on the illumination level of the scene. In a typical indoor lighting setup, it takes the prototype less than 30 seconds to acquire a 400×400×500 hyperspectral image datacube with an acceptable signal-to-noise ratio of more than 20 dB.

### 3.2. Some initial results

We examined the performance of the prototype by running extensive experiments and comparing the hyperspectral images acquired by the prototype to those obtained by a commercial VNIR hyperspectral imager supplied by Headwall Photonics [25]. The commercial device, called Headwall Hyperspec, is a line spectrometer capable of measuring fractional reflectance for 327 individual spectral bands between 400 and 1000 nm with a dynamic range of 10-bit. We utilized a moving platform to provide the lateral movement required by Headwall Hyperspec for scanning the scene to be captured.

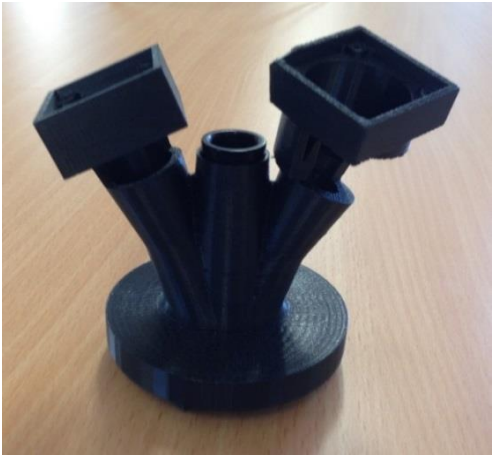
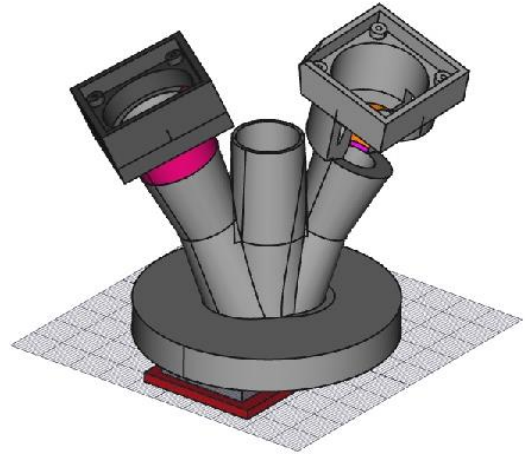

Figure 5, 3D-printed enclosure of the prototype and its CAD design. The auxiliary image sensor is placed on the left side, the spectral sensor as well as the diffraction grating on the right side, and the objective lens at the middle.

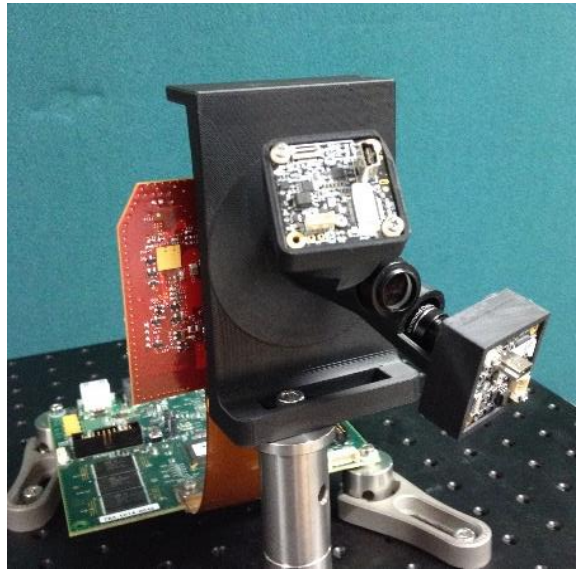

Figure 6, The prototype connected to the DMD controller.

We conducted the experiments whose results are given below within an isolated compartment illuminated by a controlled light source. However, we have also performed experiments in various indoor and outdoor environments and have found that the ambient illumination level only affects the acquisition time and not the accuracy of the prototype.

Figure 7 shows two hyperspectral image slices and their corresponding RGB images captured by the prototype during scanning a scene that contains a leaf. Each hyperspectral image slice belongs to a line of pixels noticeable as a dark stripe on the corresponding RGB image as the DMD micromirrors send only the light of those pixels to the spectral sensor. By analyzing the RGB images, we are able to locate precisely the positions within the scene that correspond to the captured spectral data. This provides an invaluable advantage over conventional pushbroom hyperspectral imagers when used on non-fixed platforms such as wheeled or airborne vehicles, which induce wobbling and vibration. Uncompensated movements can inevitably wash out the spatial resolution of the hyperspectral image captured in a scan.

Figure 8 displays parts of a hyperspectral image of the leaf shown in Figure 7 captured by the prototype. In this figure, there are 20 monochrome images each corresponding to one spectral band with a width of less than a nanometer. In Figure 9, we plot the normalized root-mean-square difference (RMSD) between the hyperspectral images acquired by the prototype and Headwall Hyperspec versus the number of spectral bands, i.e., the spectral resolution of the hyperspectral images. To obtain the results of Figure 9, we used the prototype and Headwall Hyperspec to acquire hyperspectral images of the same scene with the sensed spectrum being divided into a number of spectral bands ranging from 50 to 300. Figure 9 shows that the normalized RMSD is below 4.1% for all considered spectral resolutions confirming the accuracy of the prototype.

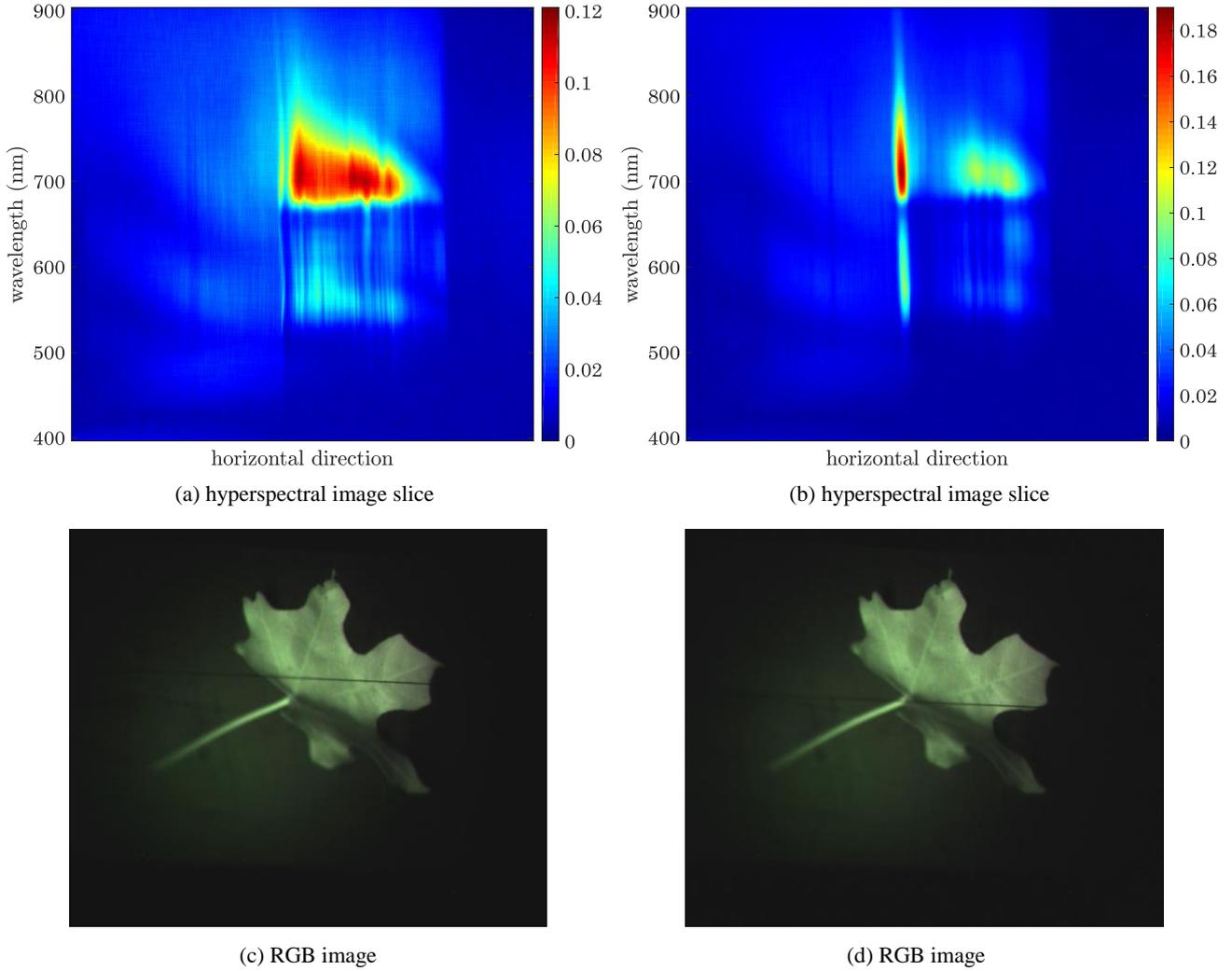

Figure 7, Two hyperspectral image slices, (a) and (b), and their corresponding RGB images, (c) and (d), respectively, captured by the prototype.

Figure 10(a) is an RGB image of a plant having three main leaves acquired by the prototype when all micromirrors on the DMD direct the incoming light towards the auxiliary image sensor. Figure 10(b) is a processed version of Figure 10(a) where the leaves have been identified and labeled using standard edge detection [26] and image segmentation tools. Figure 11 shows the average spectrum of a 16-pixel block on each leave captured by the prototype and Headwall Hyperspec. The spectra produced by the prototype are in good agreement with those produced by Headwall Hyperspec. However, it takes the prototype only a few seconds to collect the required data and produce the average spectra shown in Figure 11. It does so in an automatic manner once programmed accordingly. On the other hand, Headwall Hyperspec requires a full scan followed by manual mining of the relevant data.

## 4. CONCLUSION

We introduced a pushbroom hyperspectral imaging system that employs a digital micromirror device (DMD), a transmissive diffraction grating, and two off-the-shelf standard CMOS image sensors. One of the image sensors (spectral sensor) collects the spectral data and the other (auxiliary image sensor) takes RGB images, while the DMD takes the role of a tall narrow aperture (slit) that scans the scene. The system offers two main advantages compared with the existing hyperspectral imagers. First, it facilitates robust construction of a hyperspectral image datacube from the hyperspectral image slices collected by the spectral sensor despite the presence of possible small movements during the capture. This is because for every hyperspectral slice, an RGB image of the scene, albeit with a missing line corresponding to the location of the slit, is taken by the auxiliary image sensor. Second, it provides RGB images of the scene that are perfectly aligned with the captured hyperspectral image in the geometric sense. Therefore, it is possible to select parts of the scene by analyzing an initial RGB image and then perform the spectral imaging only on the specified regions of interest. These advantages are mostly due to the use of a DMD in place of a moving slit and the fact that both image sensors used in the design see the scene through the same objective lens. Our prototype built based on the described design demonstrates a good performance in comparison with a highly-accurate commercial hyperspectral imaging device.

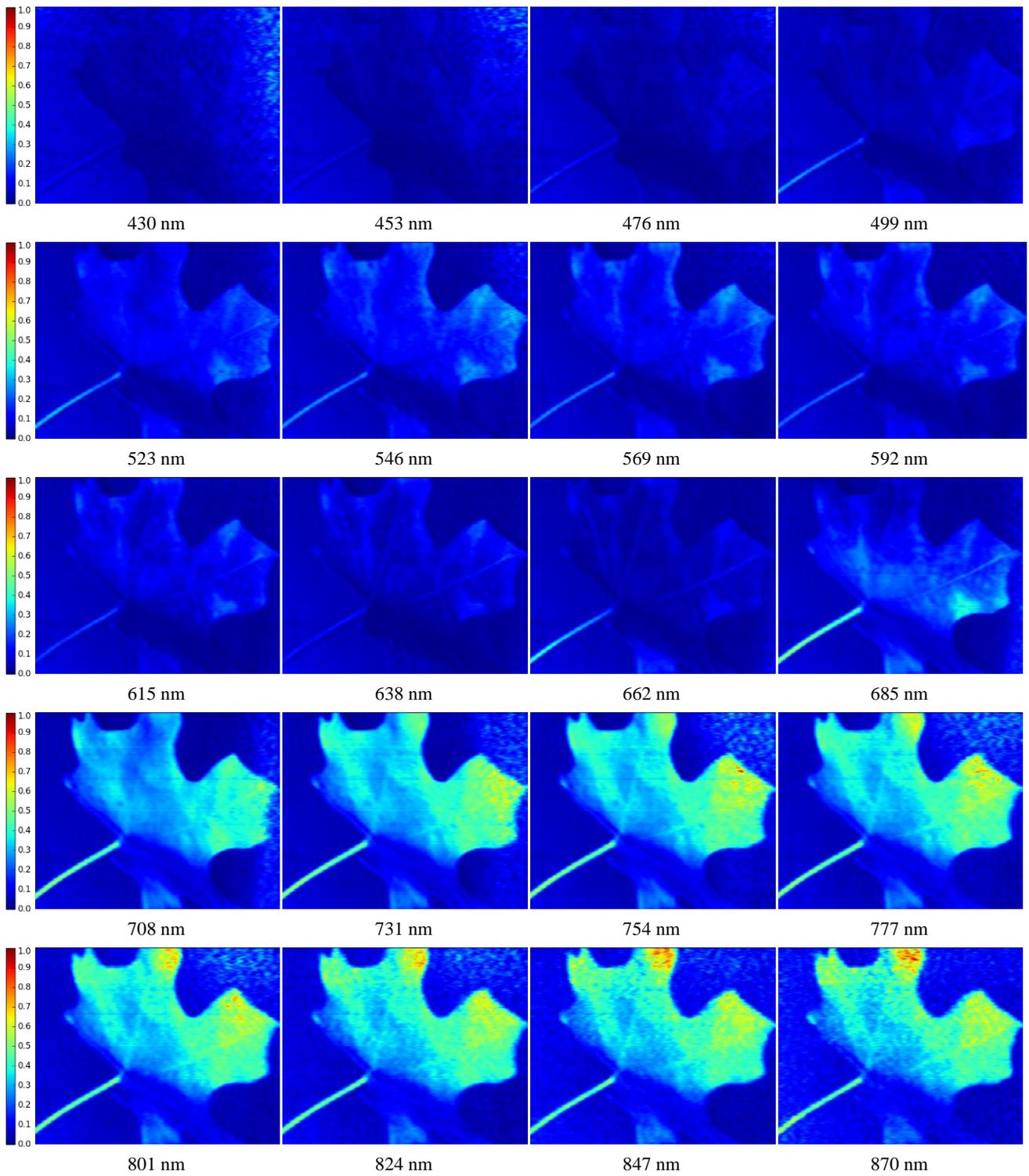

Figure 8, Images corresponding to twenty different spectral bands of a hyperspectral image captured by the prototype. The width of each spectral band is less than a nanometer.

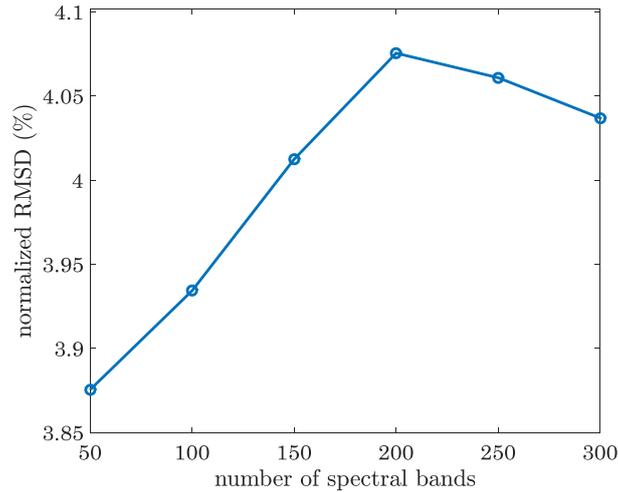

Figure 9, Normalised root-mean-square difference between the hyperspectral images acquired by the prototype and Headwall Hyperspec at various spectral resolutions.

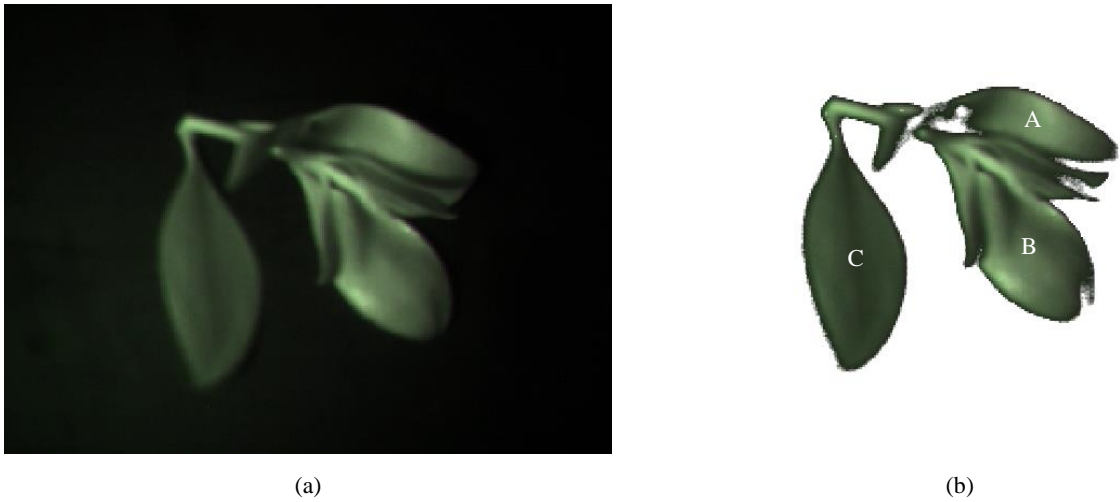

(a)            (b)

Figure 10, An RGB image captured by the auxiliary image sensor (a) and its processed version where the three main leaves are identified and labeled with A, B, and C (b).

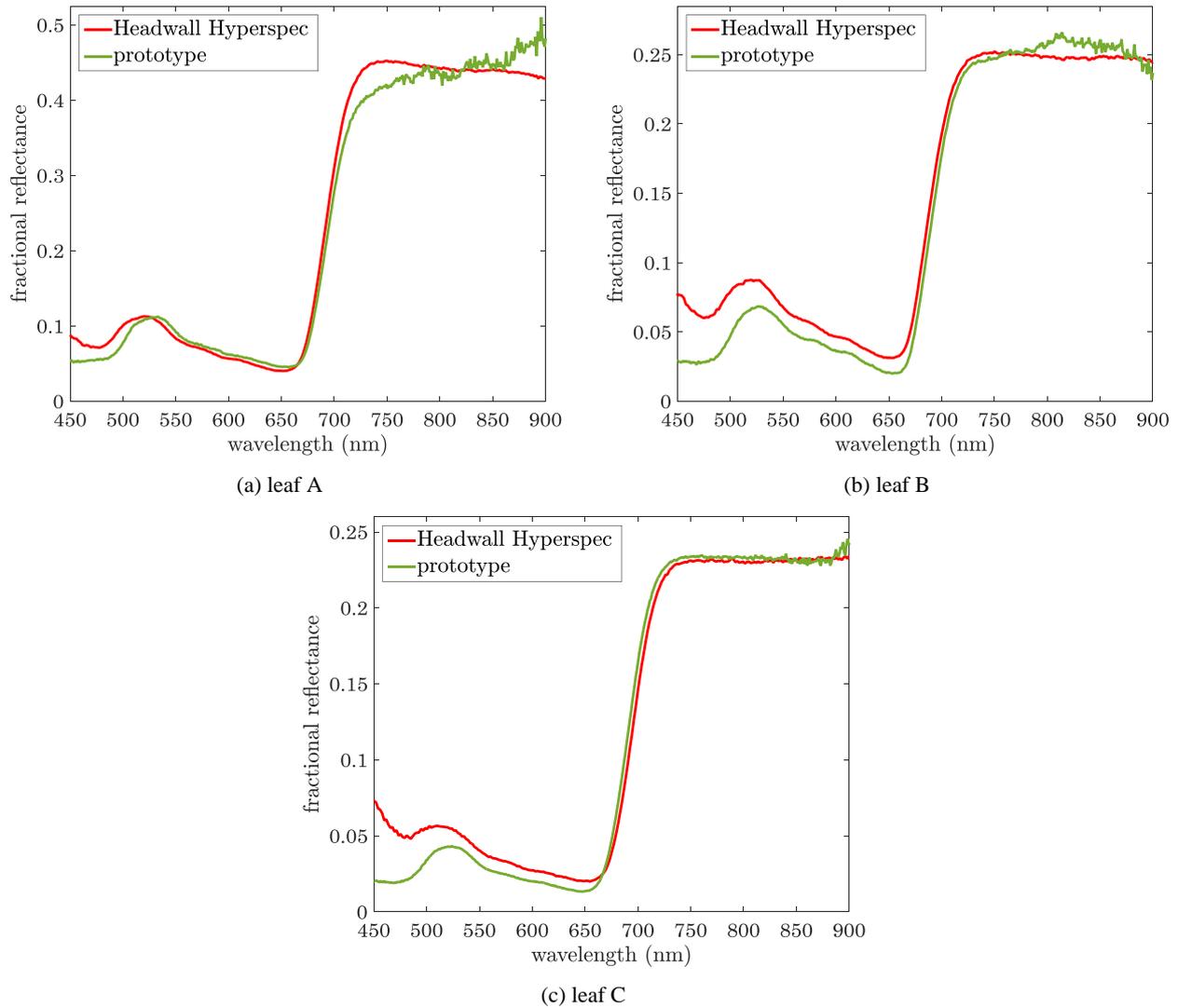

Figure 11, The average spectrum of a square block of 16 pixels on each leaf of the plant in Figure 9 captured by the prototype and Headwall Hyperspec.